\pdfoutput=1
\documentclass[11pt]{article}

\usepackage{acl}

\usepackage{times}
\usepackage{latexsym}

\usepackage[T1]{fontenc}
\usepackage[utf8]{inputenc}

\usepackage{microtype}

\usepackage{times}
\usepackage{url}
\usepackage{latexsym}

\usepackage{algorithmic}
\usepackage[linesnumbered,algoruled,boxed,lined]{algorithm2e}
\usepackage{siunitx}
\usepackage{amsmath}
\usepackage{amssymb}
\usepackage{MnSymbol}
\usepackage{color}
\usepackage{graphicx}
\usepackage{here}
\usepackage{tcolorbox}
\usepackage{threeparttable}
\usepackage{booktabs}
\usepackage{caption}
\usepackage{array}
\usepackage{hhline}
\usepackage{multirow}
\usepackage{tikz-dependency}
\usepackage{tikz-qtree}

\newdimen\algorithmwidth
\makeatletter
\edef\@algocf@start{\unexpanded{%
    \begin{center}
      \begin{minipage}{\algorithmwidth}
  }\expandafter\unexpanded\expandafter{\@algocf@start}}
\expandafter\def\expandafter\@algocf@finish\expandafter{\@algocf@finish
      \end{minipage}
    \end{center}
}
\SetAlCapNameFnt{\small}
\SetAlCapFnt{\small}

\title{Multilingual Syntax-aware Language Modeling \\ through Dependency Tree Conversion}

\author{Shunsuke Kando$^{1,2}$ \hspace{7mm} Hiroshi Noji$^{3,2}$ \hspace{7mm} Yusuke Miyao$^{1,2}$ \\
        $^1$ The University of Tokyo \\
        $^2$ Artificial Intelligence Research Center, AIST \\
        $^3$ LeapMind Inc. \\
        \texttt{kando-shunsuke@alumni.u-tokyo.ac.jp} \\
        \texttt{noji@leapmind.io} \\
        \texttt{yusuke@is.s.u-tokyo.ac.jp}
}

\begin{document}
\maketitle
\begin{abstract}
Incorporating stronger syntactic biases into neural language models (LMs) is a long-standing goal, but research in this area often focuses on modeling English text, where constituent treebanks are readily available.
Extending constituent tree-based LMs to the multilingual setting, where dependency treebanks are more common, is possible via dependency-to-constituency conversion methods.
However, this raises the question of which tree formats are best for learning the model, and for which languages.
We investigate this question by training recurrent neural network grammars (RNNGs) using various conversion methods, and evaluating them empirically in a multilingual setting.
We examine the effect on LM performance across nine conversion methods and five languages through seven types of syntactic tests.
On average, the performance of our best model represents a 19 \% increase in accuracy over the worst choice across all languages.
Our best model shows the advantage over sequential/overparameterized LMs, suggesting the positive effect of syntax injection in a multilingual setting.
Our experiments highlight the importance of choosing the right tree formalism, and provide insights into making an informed decision.
\end{abstract}

\section{Introduction}
\label{sec:introduction}
The importance of language modeling in recent years has grown considerably, as methods based on large pre-trained neural language models (LMs) have become the state-of-the-art for many problems (\citealp{devlin-etal-2019}; \citealp{radford-etal-2019}).
However, these neural LMs are based on general architectures and therefore do not explicitly model linguistic constraints, and have been shown to capture only a subset of the syntactic representations typically found in constituency treebanks \citep{warstadt-etal-2020}.
An alternative line of LM research aims to explicitly model the parse tree in order to make the LM syntax-aware.
A representative example of this paradigm, reccurent neural network grammar (RNNG, \citealp{dyer-etal-2016}), is reported to perform better than sequential LMs on tasks that require complex syntactic analysis (\citealp{kuncoro-etal-2019}; \citealp{hu-etal-2020}; \citealp{noji-oseki-2021}).

The aim of this paper is to extend LMs that inject syntax to the multilingual setting.
This attempt is important mainly in two ways.
Firstly, English has been dominant in researches on syntax-aware LM.
While multilingual LMs have received increasing attention in recent years, most of their approaches do not explicitly model syntax, such as multilingual BERT (mBERT, \citealp{devlin-etal-2019}) or XLM-R \citep{conneau-etal-2020}.
Although these models have shown high performance on some cross-lingual tasks \citep{conneau-etal-2018}, they perform poorly on a syntactic task \citep{mueller-etal-2020}.
Secondly, syntax-aware LMs have interesting features other than their high syntactic ability.
One example is the validity of RNNG as a cognitive model under an English-based setting, as demonstrated in \citet{hale-etal-2018}.
Since human cognitive functions are universal, while natural languages are diverse, it would be ideal to conduct this experiment based on multiple languages.

The main obstacle for multilingual syntax-aware modeling is that it is unclear how to inject syntactic information while training.
A straightforward approach is to make use of a multilingual treebank, such as Universal Dependencies (UD, \citealp{nivre-etal-2016}; \citealp{nivre-etal-2020}), where trees are represented in a dependency tree (DTree) formalism.
\citet{matthews-etal-2019} evaluated parsing and language modeling performance on three typologically different languages, using a generative dependency model.
Unfortunately, they revealed that dependency-based models are less suited to language modeling than comparable constituency-based models, highlighting the apparent difficulty of extending syntax-aware LMs to other languages using existing resources.

This paper revisits the issue of the difficulty of constructing multilingual syntax-aware LMs, by exploring the performance of multilingual language modeling using constituency-based models.
Since our domain is a multilingual setting, our focus turns to how dependency-to-constituency conversion techniques result in different trees, and how these trees affect the model's performance.
We obtain constituency treebanks from UD-formatted dependency treebanks of five languages using nine tree conversion methods.
These treebanks are in turn used to train an RNNG, which we evaluate on perplexity and CLAMS \citep{mueller-etal-2020}.

Our contributions are: (1) We propose a methodology for training multilingual syntax-aware LMs through the dependency tree conversion. (2) We found an optimal structure that brings out the potential of RNNG across five languages. (3) We demonstrated the advantage of our multilingual RNNG over sequential/overparameterized LMs.

\section{Background}
\label{sec:background}

\subsection{Recurrent Neural Network Grammars}
\label{ssec:rnng}
\begin{figure}
\centering
\setlength{\tabcolsep}{1mm}
\begin{tabular}{c}
\begin{tabular}{l|l|l}
\toprule
\begin{tabular}{l}
\tiny
\tikzset{level distance=12pt}
\Tree [.S
        [.NP \edge[roof]; {The pilot} ]
        [.VP laughs ]
      ]
\end{tabular}
&
\begin{tabular}{l}
{\small\bf Partial tree} \\
{\small\bf Stack-LSTM}
\end{tabular}
& {\small\bf Action} \\\hline
{\tiny 0} &
& \small NT(S) \\\hline
{\tiny 1} &
\small
\begin{tabular}{l}
(S \\
$[\mathbf{e}_{\rm S}]$
\end{tabular} & \small NT(NP) \\\hline
{\tiny 2} &
\small
\begin{tabular}{l}
(S (NP \\
$[\mathbf{e}_{\rm S} \ \mathbf{e}_{\rm NP}]$
\end{tabular} & \small GEN(The) \\\hline
{\tiny 3} &
\small
\begin{tabular}{l}
(S (NP The \\
$[\mathbf{e}_{\rm S} \ \mathbf{e}_{\rm NP} \ \mathbf{e}_{\rm The}]$
\end{tabular} & \small GEN(pilot) \\\hline
{\tiny 4} &
\small
\begin{tabular}{l}
(S (NP The pilot \\
$[\mathbf{e}_{\rm S} \ \underline{\mathbf{e}_{\rm NP} \ \mathbf{e}_{\rm The} \ \mathbf{e}_{\rm pilot}}]$
\end{tabular} & \small REDUCE \\\hline
{\tiny 5} &
\small
\begin{tabular}{l}
(S (NP The pilot) \\
$[\mathbf{e}_{\rm S} \ \underline{\mathbf{e}_{\rm NP'}}]$
\end{tabular} & \small NT(VP) \\\hline
{\tiny 6} &
\small
\begin{tabular}{l}
(S (NP The pilot) (VP \\
$[\mathbf{e}_{\rm S} \ \mathbf{e}_{\rm NP'} \ \mathbf{e}_{\rm VP}]$
\end{tabular} & \small $\cdots$ \\
\bottomrule
\end{tabular}
\end{tabular}
\caption{The illustration of stack-RNNG behavior.
Stack-LSTM represents the current partial tree, in which adjacent vectors are connected in the network.
At REDUCE action, the corresponding vector is updated with composition function (as underlined).
}
\label{fig:stackRNNG}
\end{figure}

RNNGs are generative models that estimate joint probability of a sentence $\boldsymbol{x}$ and a constituency tree (CTree) $\boldsymbol{y}$.
The probability $p(\boldsymbol{x}, \boldsymbol{y})$ is estimated with top-down constituency parsing actions $\boldsymbol{a} = (a_1, a_2, \cdots, a_n)$ that produce $\boldsymbol{y}$:

\[
p(\boldsymbol{x}, \boldsymbol{y}) = \prod_{t=1}^n p(a_t | a_1, \cdots, a_{t-1})
\]

\citet{kuncoro-etal-2017} proposed a stack-only RNNG that computes the next action probability based on the current partial tree.
Figure~\ref{fig:stackRNNG} illustrates the behavior of it.
The model represents the current partial tree with a stack-LSTM, which consists of three types of embeddings: nonterminal, word, and closed-nonterminal. 
The next action is estimated with the last hidden state of a stack-LSTM.
There are three types of actions as follows:
\begin{itemize}
\item NT(X): Push nonterminal embedding of $X$ ($\mathbf{e}_{X}$) onto the stack.
\item GEN($w$): Push word embedding of $w$ ($\mathbf{e}_{w}$) onto the stack.
\item REDUCE: Pop elements from the stack until a nonterminal embedding shows up. With all the embeddings which are popped, compute closed-nonterminal embedding $\mathbf{e}_{X'}$ using composition funcion \textsc{Comp}:
\[
\mathbf{e}_{X'} = \textsc{Comp}(\mathbf{e}_{X}, \mathbf{e}_{w_1}, \cdots, \mathbf{e}_{w_m})
\]
\end{itemize}

RNNG can be regarded as a language model that injects syntactic knowledge explicitly, and various appealing features have been reported (\citealp{kuncoro-etal-2017}; \citealp{kuncoro-etal-2017}; \citealp{hale-etal-2018}).
We focus on its high performance on \emph{syntactic evaluation}, which is described below.

\paragraph{Difficulty in extending to other languages}
In principle, RNNG can be learned with any corpus as long as it contains CTree annotation.
However, it is not evident which tree formats are best in a multilingual setting.
Using the same technique as English can be inappropriate because each language has its own characteristic, which can be different from English.
This question is the fundamental motivation of this research.

\subsection{Cross-linguistic Syntactic Evaluation}
\label{ssec:CLAMS}
To investigate the capability of LMs to capture syntax, previous work has attempted to create an evaluation set that requires analysis of the sentence structure \citep{linzen-etal-2016}. 
One typical example is a subject-verb agreement, a rule that the form of a verb is determined by the grammatical category of the subject, such as person or number:
\begin{equation}
\label{eq:orc}
\text{The pilot that the guards love laughs/*laugh.}
\end{equation}

In (\ref{eq:orc}), the form of \emph{laugh} is determined by the subject \emph{pilot}, not \emph{guards}.
This judgment requires syntactic analysis; \emph{guards} is not a subject of target verb \emph{laugh} because it is in the relative clause of the real subject \emph{pilot}.

\citet{marvin-linzen-2018} designed the English evaluation set using a grammatical framework.
\citet{mueller-etal-2020} extended this framework to other languages (French, German, Hebrew, and Russian) and created an evaluation set named CLAMS (Cross-Linguistic Assessment of Models on Syntax).
CLAMS covers 7 categories of agreement tasks, including local agreement (e.g. The author \underline{laughs/*laugh}) and non-local agreement that contains an intervening phrase between subject and verb as in (\ref{eq:orc}).
They evaluated LMs on CLAMS and demonstrated that sequential LMs often fail to assign a higher probability to the grammatical sentence in cases that involve non-local dependency.

Previous work has attempted to explore the syntactic capabilities of LMs with these evaluation sets.
\citet{kuncoro-etal-2019} compared the performance of LSTM LM and RNNG using the evaluation set proposed in \citet{marvin-linzen-2018}, demonstrating the superiority of RNNG in predicting the agreement.
\citet{noji-takamura-2020} suggested that LSTM LMs potentially have a limitation in handling object relative clauses.
Since these analyses are performed on the basis of English text, it is unclear whether they hold or not in a multilingual setting.
In this paper, we attempt to investigate this point by learning RNNGs in other languages and evaluating them on CLAMS.

\begin{algorithm}[t]
  \small
  \caption{\texttt{lf} is short for left-first conversion.
  We omit right-first conversion because it can be defined just by swapping the codeblocks 6-9 and 10-13 of left-first conversion.}
  \label{algo:conv}
  \SetKw{KwTo}{in} 
  \SetKw{for}{for} 
  \SetKw{Continue}{continue}
  \SetKw{Return}{return}
  \SetKwFunction{removeEmptyList}{removeEmptyList}
  \SetKwFunction{ldeps}{ldeps}
  \SetKwFunction{rdeps}{rdeps}
  \SetKwFunction{pop}{pop}
  \SetKwFunction{flat}{flat}
  \SetKwFunction{lf}{lf}
  \SetKwProg{Fn}{Function}{:}{}
  \Fn{\flat{$w$, $ldeps$, $rdeps$}}{
    lNT $\leftarrow$ [\flat{$lw$, $lw$.\ldeps, $lw$.\rdeps} \for $lw$ \KwTo $ldeps$]\;
    rNT $\leftarrow$ [\flat{$rw$, $rw$.\ldeps, $rw$.\rdeps} \for $rw$ \KwTo $rdeps$]\;
    \Return [lNT [$w$] rNT].\removeEmptyList \;
  }
  \Fn{\lf{$w$, $ldeps$, $rdeps$}}{
    \uIf{$ldeps$ {\rm is not empty}}{
      \tcc{Pop left-most dependent}
      $lw$ $\leftarrow$ $ldeps$.\pop{}\;
      lNT $\leftarrow$ [\lf{$lw$, $lw$.\ldeps, $lw$.\rdeps}]\;
      rNT $\leftarrow$ [\lf{$w$, $ldeps$, $rdeps$}]\;
    }
    \uElseIf{$rdeps$ {\rm is not empty}}{
      \tcc{Pop right-most dependent}
      $rw$ $\leftarrow$ $rdeps$.\pop{}\;
      lNT $\leftarrow$ [\lf{$w$, $ldeps$, $rdeps$}]\;
      rNT $\leftarrow$ [\lf{$rw$, $rw$.\ldeps, $rw$.\rdeps}]\;
    }
    \lElse{\Return [$w$]}
    \Return [lNT rNT]\;
  }
    % \uIf{\rm done[node] or node.dependent = []}{
    %   node.children $\leftarrow$ []\;
    %   \Return node\;
    % }
    % \rm children $\leftarrow$ \concat([node], node.dependent)\;
    % \tcc{let children be a list including {\rm node} and its dependents}
    % %\tcp*[r]{this is comment}
    % \uIf{\rm method $=$ flat}{
    %   node.children $\leftarrow$ children\;
    %   done[node] $\leftarrow$ true\;
    %   \lFor{\rm child \KwTo children.fromleft}{
    %     \convert(child)
    %   }
    % }
    % \uElseIf{\rm method $=$ left-first}{
    %   \For(\tcp*[f]{loop from the left most child}){\rm child \KwTo children.fromleft}{
    %     \lIf{\rm child.id $<$ node.id}{
    %       node.children $\leftarrow$ [child, node]
    %     }
    %     \lElseIf{\rm child.id $>$ node.id}{
    %       node.children $\leftarrow$ [node, child]
    %     }
    %     \lElse{
    %       \Continue
    %     }
    %     done[node] $\leftarrow$ true; 
    %     \convert{\rm child};
    %     \convert{\rm node}; 
    %   }
    % }
    % \uElseIf(\tcp*[f]{loop from the right most child}){\rm method $=$ right-first}{
    %   \lFor{\rm child \KwTo children.fromright}{
    %     \emph{the same as line 9 - 13.}
    %   }
    % }
    % \Return node;
\end{algorithm}

\section{Method: Dependency Tree Conversion}
\label{sec:method}
As a source of multilingual syntactic information, we use Universal Dependencies (UD), a collection of cross-linguistic dependency treebanks with a consistent annotation scheme.
Since RNNG requires a CTree-formatted dataset for training, we perform DTree-to-CTree conversions, which are completely algorithmic to make it work regardless of language.
Our method consists of two procedures: \textbf{structural conversion} and \textbf{nonterminal labeling}; obtaining a CTree skeleton with unlabeled nonterminal nodes, then assigning labels by leveraging syntactic information contained in the dependency annotations.
While our structural conversion is identical to the baseline approach of \citet{collins-etal-1999}, we include a novel labeling method that relies on dependency relations, not POS tags.

% Most work on DTree-to-CTree conversion is a supervised approach that requires both dependency and constituency annotations (\citealp{kong-etal-2015}; \citealp{lee-wang-2016}).
% Since these methods do not apply to our case, we adopted a completely algorithmic approach.
% While we used the baseline approach of \citet{collins-etal-1999} for structural conversion, we include a novel labeling method that relies on dependency label, not POS tag.

\begin{figure*}[t]
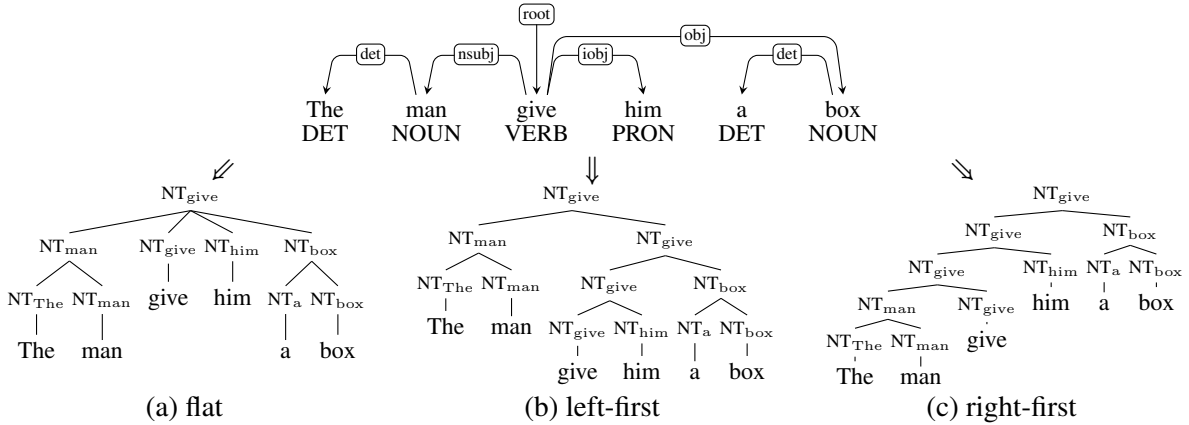

  \centering
  \begin{tabular}{ccc}
    \multicolumn{3}{c}{
    \begin{dependency}
      \footnotesize
        \begin{deptext}[column sep=3ex]
        The \& man \& give \& him \& a \& box \\
        DET \& NOUN \& VERB \& PRON \& DET \& NOUN \\
        \end{deptext}
        \deproot[edge unit distance=2.2ex]{3}{root}
        \depedge{2}{1}{det}
        \depedge{3}{2}{nsubj}
        \depedge{3}{4}{iobj}
        \depedge[edge unit distance=1.8ex]{3}{6}{obj}
        \depedge{6}{5}{det}
    \end{dependency}
    }
    \vspace{-2mm}
    \\
    \hspace{1cm}$\Swarrow$\vspace{-2mm} &
    $\Downarrow$ &
    \hspace{-1cm}$\Searrow$
    \\
    \scriptsize
    \tikzset{level distance=20pt, sibling distance=-2pt}
    \Tree [.NT$_{\mathrm{give}}$ 
        [.NT$_{\mathrm{man}}$
          [.NT$_{\mathrm{The}}$ {\small The} ]
          [.NT$_{\mathrm{man}}$ {\small man} ]
        ]
        [.NT$_{\mathrm{give}}$ {\small give} ]
        [.NT$_{\mathrm{him}}$ {\small him} ]
        [.NT$_{\mathrm{box}}$
          [.NT$_{\mathrm{a}}$ {\small a} ]
          [.NT$_{\mathrm{box}}$ {\small box} ]
        ]
      ]
    &
    \tikzset{level distance=17pt, sibling distance=-2pt}
    \scriptsize
    \Tree [.NT$_{\mathrm{give}}$ 
        [.NT$_{\mathrm{man}}$
          [.NT$_{\mathrm{The}}$ {\small The} ]
          [.NT$_{\mathrm{man}}$ {\small man} ]
        ]
        [.NT$_{\mathrm{give}}$
          [.NT$_{\mathrm{give}}$
            [.NT$_{\mathrm{give}}$ {\small give} ]
            [.NT$_{\mathrm{him}}$ {\small him} ]
          ]
          [.NT$_{\mathrm{box}}$
            [.NT$_{\mathrm{a}}$ {\small a} ]
            [.NT$_{\mathrm{box}}$ {\small box} ]
          ]
        ]
      ]
    &
    \tikzset{level distance=14pt, sibling distance=-2pt}
    \scriptsize
    \Tree [.NT$_{\mathrm{give}}$ 
        [.NT$_{\mathrm{give}}$
          [.NT$_{\mathrm{give}}$
            [.NT$_{\mathrm{man}}$
              [.NT$_{\mathrm{The}}$ {\small The} ]
              [.NT$_{\mathrm{man}}$ {\small man} ]
            ]
            [.NT$_{\mathrm{give}}$ {\small give} ]
          ]
          [.NT$_{\mathrm{him}}$ {\small him} ]
        ]
        [.NT$_{\mathrm{box}}$
          [.NT$_{\mathrm{a}}$ {\small a} ]
          [.NT$_{\mathrm{box}}$ {\small box} ]
        ]
      ]
    \\
    (a) flat & (b) left-first & (c) right-first
  \end{tabular}
  \caption{The illustration of structural conversion. NT$_w$ is a temporal label of nonterminal which will be assigned at nonterminal labeling phase.}
  \label{fig:conversion}
\end{figure*}

\paragraph{Structural conversion}
We performed three types of structural conversion: \emph{flat}, \emph{left-first}, and \emph{right-first}.
Algorithm~\ref{algo:conv} shows the pseudo code and Figure~\ref{fig:conversion} illustrates the actual conversions.
These approaches construct CTree in a top-down manner following this procedure: 1) Introduce the root nonterminal of the head of a sentence (NT$_\mathrm{give}$). 2) For each NT$_w$, introduce new nonterminals according to the dependent(s) of $w$. Repeat this procedure recursively until $w$ has no dependents.

The difference between the three approaches is the ordering of introducing nonterminals.
We describe their behaviors based on the example in Figure~\ref{fig:conversion}.
(a) flat approach lets $w$ and its dependents be children in CTree simultaneously.
For example, NT$_{\mathrm{give}}$ has four children: NT$_{\mathrm{man}}$, NT$_{\mathrm{give}}$, NT$_{\mathrm{him}}$, NT$_{\mathrm{box}}$, because they are dependents of the head word \emph{give}.
As the name suggests, this approach tends to produce a flat-structured CTree because each nonterminal can have multiple children.
(b) left-first approach introduces the nonterminals from the left-most dependent.
If there is no left dependent, the right-most dependent is introduced.
In the example of Figure~\ref{fig:conversion}, the root NT$_\mathrm{give}$ has a left child NT$_\mathrm{man}$ because \emph{man} is the left-most dependent of the head \emph{give}.
(c) right-first approach is the inversed version of left-first; handling the right-most dependent first.
For methods (b) and (c), the resulting CTree is always a binary tree.

\begin{table}
\centering
\small
\begin{tabular}{lccc}
\hline
                        & X-label & POS-label & DEP-label \\\hline
NT$_{\mathrm{The}}$     & X       & DETP      & det       \\
NT$_{\mathrm{man}}$     & X       & NOUNP     & nsubj     \\
NT$_{\mathrm{give}}$    & X       & VERBP     & root      \\
NT$_{\mathrm{him}}$     & X       & PRONP     & iobj      \\
NT$_{\mathrm{a}}$       & X       & DETP      & det       \\
NT$_{\mathrm{box}}$     & X       & NOUNP     & obj       \\
\hline
\end{tabular}
\caption{Actual labels assigned to nonterminals.}
\label{tab:label}
\end{table}

\paragraph{Nonterminal labeling}
We define three types of labeling methods for each NT$_{w}$; 1) X-label: Assign ``X'' to all the nonterminals. 2)POS-label: Assign POS tag of $w$. 3) DEP-label: Assign dependency relation between $w$ and its head.
Table~\ref{tab:label} shows the actual labels that are assigned to CTrees in Figure~\ref{fig:conversion}.

\tikzset{baseline,every tree node/.style={align=center,anchor=north}} 
\begin{figure*}
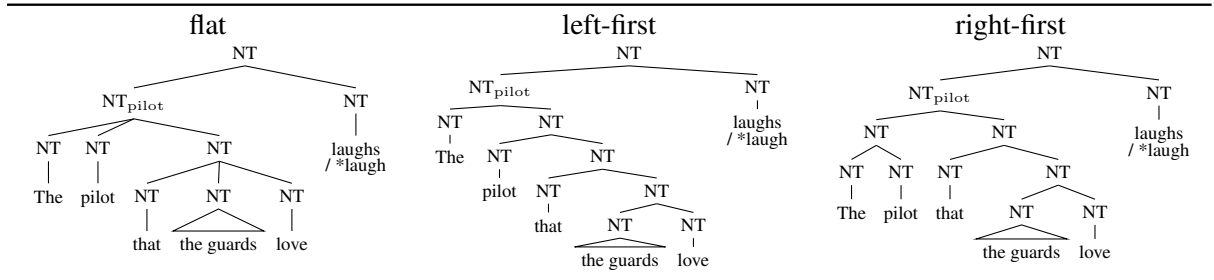

  \scriptsize
  \begin{tabular}{ccc}
    \toprule
    \normalsize
    flat\vspace{-2mm} &
    \normalsize
    left-first &
    \normalsize
    right-first \\
    \tikzset{level distance=18pt}
    \Tree [.NT
            [.NT$_\mathrm{pilot}$
              [.NT The ]
              [.NT pilot ]
              [.NT
                [.NT that ]
                [.NT \edge[roof]; {the guards} ]
                [.NT love ]
              ]
            ]
            [.NT {laughs \\ / *laugh} ]
          ]
    &
    \tikzset{level distance=13pt}
    \Tree [.NT
        [.NT$_\mathrm{pilot}$
          [.NT The ]
          [.NT
            [.NT pilot ]
            [.NT
              [.NT that ]
              [.NT
                [.NT \edge[roof]; {the guards} ]
                [.NT love ]
              ]
            ]
          ]
        ]
        [.NT {laughs \\ / *laugh} ]
      ] 
    &
    \tikzset{level distance=15pt}
    \Tree [.NT
      [.NT$_\mathrm{pilot}$
        [.NT
          [.NT The ]
          [.NT pilot ]
        ]
        [.NT
          [.NT that ]
          [.NT
            [.NT \edge[roof]; {the guards} ]
            [.NT love ]
          ]
        ]
      ]
      [.NT {laughs \\ / *laugh} ]
    ]
    \\
    \bottomrule
  \end{tabular}
  \caption{Examples of converted CTrees. A sentence is taken from CLAMS, which requires recognition of long distance dependency intervened by object relative clause (sentence (\ref{eq:orc})). For simplicity, we omit the corresponding word of each nonterminal except for \emph{pilot}, the main subject of the sentence.}
  \label{fig:clams-structure}
\end{figure*}

Each method has its own intent.
X-label drops the syntactic category of each phrase, which minimizes the structural information of the sentence.
POS-label would produce the most common CTree structure because traditionally nonterminals are labeled based on POS tag of the head word.
DEP-label is a more fine-grained method than POS-label because words in a sentence can have the same POS tag but different dependency relation, as in \emph{man} and \emph{box} in Figure~\ref{fig:conversion}.

Finally, we performed a total of nine types of conversions (three structures $\times$ three labelings).
Although they have discrete features, they are common in that they embody reasonable phrase structures that are useful for capturing syntax.
Figure~\ref{fig:clams-structure} shows the converted structure of an actual instance from CLAMS.
In all settings, the main subject phrase is correctly dominated by NT$_\mathrm{pilot}$, which should contribute to solving the task.

\section{What Is the Robust Conversion Which Works Well in Every Language?}
\label{sec:compare}
In Section~\ref{sec:method}, we proposed language-independent multiple conversions from DTree to CTree.
The intriguing question is; Is there a robust conversion that brings out the potential of RNNG in every language?
To answer this question, we conducted a thorough experiment to compare the performances of RNNGs trained in each setting.

\subsection{Experimental Setup}
\label{ssec:compare-exp}
\paragraph{Treebank preparation}
Following \citet{mueller-etal-2020}, we extracted Wikipedia articles of target languages using WikiExtractor\footnote{\url{https://github.com/attardi/ wikiextractor}} to create corpora\footnote{Although \citet{mueller-etal-2020} publishes corpora they used, we extracted the dataset ourselves because they contain \texttt{<unk>} token which would affect parsing.}.
We fed it to UDify \citep{kondratyuk-straka-2019}, a multilingual neural dependency parser trained on the entire UD treebanks, to generate a CoNLL-U formatted dependency treebank.
Sentences are tokenized beforehand using Stanza \citep{qi-etal-2020} because UDify requires tokenized text for prediction.
The resulting dependency treebank is converted into the constituency treebank using methods proposed in Section~\ref{sec:method}.
Our treebank contains around 10\% non-projective DTrees for all the language (between 9\% in Russian and 14\% in Hebrew), and we omit them in the conversion phase because we cannot obtain valid CTrees from them\footnote{Since other language can contain more non-projective DTrees, we have to consider how to handle it in the future.}.
As a training set, we picked sentences with 10M tokens at random for each language. 
For a validation and a test set, we picked 5,000 sentences respectively.

\paragraph{Training details}
We used batched RNNG \citep{noji-oseki-2021} to speed up our training.
Following \citet{noji-oseki-2021}, we used subword units \citep{sennrich-etal-2016} with a vocabulary size of 30K.
We set the hyperparameters so as to make the model size 35M.
We trained each model for 24 hours on a single GPU.

\paragraph{Evaluation metrics}
To compare the performance among conversions, we evaluated the model trained on each dataset in two aspects: \textbf{perplexity} and \textbf{syntactic ability} based on CLAMS.

Perplexity is a standard metric for assessing the quality of LM.
Since we adopt subword units, we regard a word probability as a product of its subwords' probabilities.
To compute it on RNNG, we performed word-synchronous beam search \citep{stern-etal-2017}, a default approach implemented in batched RNNG.
Following \citet{noji-oseki-2021}, we set a beam size $k$ as 100, a word beam size $k_w$ as 10, and fast-track candidates $k_s$ as 1.
Syntactic ability is assessed by accuracy on CLAMS, which is calculated by comparing the probabilities assigned to a grammatical and an ungrammatical sentence.
If the model assigns a higher probability to a grammatical sentence, then we regard it as correct.
Chance accuracy is 0.5.

We run the experiment three times with different random seeds for initialization of the model, and report the average score with standard deviation.

\begin{figure*}[t]
  \includegraphics[width=\linewidth]{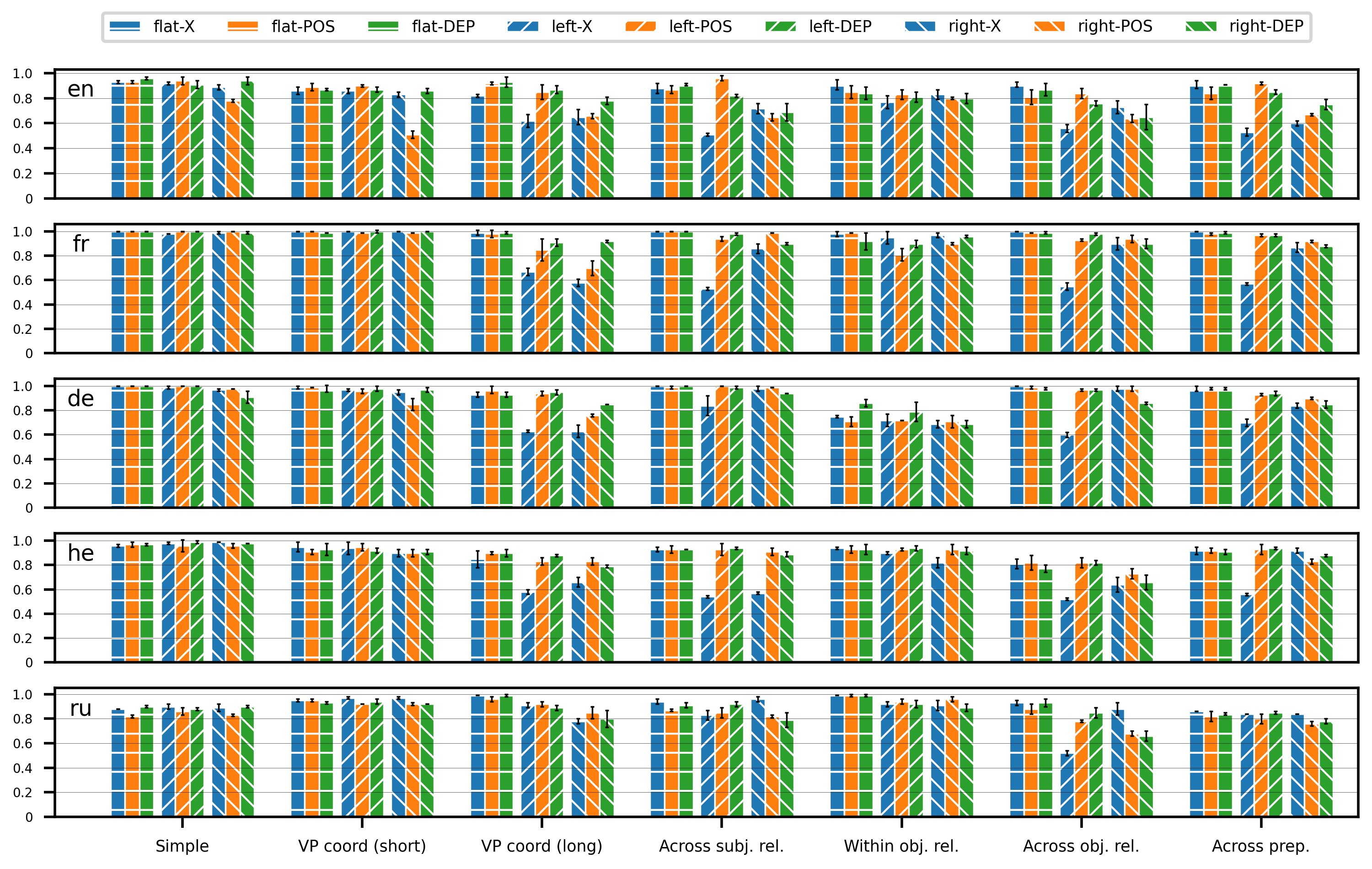}
  \caption{Accuracies of CLAMS for RNNGs trained on each setting.}
  \label{fig:10M-clams}
\end{figure*}

\subsection{Result}
\label{ssec:10M-result}
From now on, we refer to each conversion method according to a naming of the procedure, such as ``left-first structure'' or ``flat-POS conversion''.

\begin{table}[t]
\centering
\small
\begin{tabular}{c|rrr||c}
\hline
    & \multicolumn{1}{c}{flat}
    & \multicolumn{1}{c}{left}
    & \multicolumn{1}{c||}{right}
    & \\\hline
X   & 259$_{\pm1}$ & 707$_{\pm19}$ & 1507$_{\pm14}$ & \multirow{3}{*}{English}\\
POS & 278$_{\pm3}$ & 417$_{\pm2}$ & 512$_{\pm3}$  & \\
DEP & \bf{241}$_{\pm30}$ & 390$_{\pm4}$  & 463$_{\pm1}$  & \\\hline
X   & 133$_{\pm0}$ & 405$_{\pm10}$ & 691$_{\pm10}$ & \multirow{3}{*}{French}\\
POS & \bf{129}$_{\pm1}$ & 206$_{\pm2}$ & 262$_{\pm1}$ & \\
DEP & 137$_{\pm22}$ & 190$_{\pm5}$ & 223$_{\pm2}$ & \\\hline
X   & 341$_{\pm1}$ & 830$_{\pm8}$  & 1124$_{\pm18}$& \multirow{3}{*}{German}\\
POS & 366$_{\pm1}$ & 321$_{\pm3}$ & 482$_{\pm2}$  & \\
DEP & 330$_{\pm43}$ & \bf{291}$_{\pm3}$ & 398$_{\pm4}$  & \\\hline
X   & 100$_{\pm1}$ & 294$_{\pm3}$ & 450$_{\pm8}$ & \multirow{3}{*}{Hebrew}\\
POS & 97$_{\pm0}$ & 153$_{\pm1}$ & 183$_{\pm1}$ & \\
DEP & \bf{93}$_{\pm1}$ & 143$_{\pm1}$ & 161$_{\pm1}$ & \\\hline
X   & 508$_{\pm5}$ & 1413$_{\pm16}$  & 1910$_{\pm59}$& \multirow{3}{*}{Russian}\\
POS & 527$_{\pm3}$ & 845$_{\pm2}$ & 1067$_{\pm16}$& \\
DEP & \bf{473}$_{\pm61}$ & 834$_{\pm5}$ & 1030$_{\pm27}$& \\
\hline
\end{tabular}
\caption{Test set perplexity of each setting. Lower is better. ``left'' and ``right'' in the table are abbreviations of ``left-first'' and ``right-first'', respectively.}
\label{tab:10M-ppl}
\end{table}
\begin{table}
\centering
\small
\begin{tabular}{c|rrr||c}
\hline
    & \multicolumn{1}{c}{flat}
    & \multicolumn{1}{c}{left}
    & \multicolumn{1}{c||}{right}
    & \\\hline
X   & 0.89$_{\pm.01}$ & 0.68$_{\pm.01}$ & 0.75$_{\pm.01}$ & \multirow{3}{*}{English}\\
POS & 0.87$_{\pm.02}$ & 0.89$_{\pm.01}$ & \emph{0.67}$_{\pm.01}$ & \\
DEP & \bf{0.90}$_{\pm.02}$ & 0.84$_{\pm.01}$ & 0.78$_{\pm.04}$ & \\\hline
X   & \bf{0.99}$_{\pm.00}$ & \emph{0.75}$_{\pm.00}$ & 0.88$_{\pm.02}$ & \multirow{3}{*}{French}\\
POS & \bf{0.99}$_{\pm.00}$ & 0.93$_{\pm.02}$ & 0.92$_{\pm.01}$ & \\
DEP & 0.98$_{\pm.01}$ & 0.96$_{\pm.01}$ & 0.94$_{\pm.01}$ & \\\hline
X   & 0.95$_{\pm.00}$ & \emph{0.78}$_{\pm.01}$ & 0.86$_{\pm.01}$ & \multirow{3}{*}{German}\\
POS & 0.95$_{\pm.01}$ & 0.93$_{\pm.01}$ & 0.88$_{\pm.01}$ & \\
DEP & \bf{0.96}$_{\pm.01}$ & 0.95$_{\pm.02}$ & 0.87$_{\pm.02}$ & \\\hline
X   & 0.91$_{\pm.01}$ & \emph{0.72}$_{\pm.01}$ & 0.78$_{\pm.01}$ & \multirow{3}{*}{Hebrew}\\
POS & 0.91$_{\pm.01}$ & 0.91$_{\pm.03}$ & 0.87$_{\pm.01}$ & \\
DEP & 0.90$_{\pm.01}$ & \bf{0.92}$_{\pm.00}$ & 0.86$_{\pm.01}$ & \\\hline
X   & \bf{0.93}$_{\pm.00}$ & 0.84$_{\pm.01}$ & 0.89$_{\pm.02}$ & \multirow{3}{*}{Russian}\\
POS & 0.90$_{\pm.01}$ & 0.87$_{\pm.01}$ & 0.83$_{\pm.01}$ & \\
DEP & \bf{0.93}$_{\pm.01}$ & 0.89$_{\pm.00}$ & \emph{0.82}$_{\pm.01}$ & \\
\hline
\end{tabular}
\caption{CLAMS scores averaged by task category.}
\label{tab:10M-clams}
\end{table}

\paragraph{Perplexity}
Table~\ref{tab:10M-ppl} shows the perplexities in each setting.
As a whole, flat structures show the lowest perplexity, followed by left-first and right-first, which is consistent across languages.
While flat structure produces stable and relatively low perplexity regardless of labeling methods and languages, left-first and right-first structures perform very poorly on X-label.

\paragraph{Syntactic ability}
Figure~\ref{fig:10M-clams} shows the accuracies of CLAMS in each setting, and Table~\ref{tab:10M-clams} shows the average scores.
From Table~\ref{tab:10M-clams}, we observe clear distinctions across methods; the best model (shown in bold) is 19\% more accurate in average than the worst one (shown in italic), across all languages, indicating the model's certain preference for the structure.
Similar to perplexity, flat structure performs better and more stably than the others, regardless of labels and languages.
While \citet{mueller-etal-2020} reported a high variability in scores across languages when an LSTM LM is used, flat structure-based RNNGs do not show such a tendency; almost all the accuracies are above 90\%.

Looking closely at the Figure~\ref{fig:10M-clams}, we can see that left-first and right-first structures exhibit unstable behavior depending on the labeling; the accuracy on X-label tends to be lower especially for the categories that require the resolution of a long-distance dependency, such as `VP coord (long)', `Across subj. rel.', `Across obj. rel.', and `Across prep'.

\paragraph{Discussion}
Basically, we observed a similar tendency in perplexity and CLAMS score; (1) flat structures show the highest scores. (2) left-first and right-first structures perform poorly on X-label.
We conjecture that these tendencies are due to the resulting structure of each conversion; while flat structure is non-binary, the rest two are binary.
Since nonterminals in a non-binary tree can have multiple words as children, parsing actions obtained from it contain more continuous GEN actions than a binary tree.
This nature helps the model to predict the next word by considering lexical relations, which would contribute to its lower perplexity.
Although binary trees get better with the hint of informative labels (POS/DEP), it is difficult to reach the performance of flat structures due to their confused actions; GEN actions tend to be interrupted by other actions.
Besides, there are too many NT actions in a binary tree, which can hurt the prediction because the information of an important nonterminal (e.g. NT$_\mathrm{pilot}$ in Figure~\ref{fig:clams-structure}) can be diluted through the actions.
The situation becomes worse on X-label; the model cannot distinguish the nonterminal of the main subject and that of the other, resulting in missing what the subject is.

It is worth noting that perplexity does not always reflect the CLAMS accuracy.
For example, while right-X conversion produces the worst perplexity for all the languages, it achieves better CLAMS accuracy than left-X conversion for almost all the cases.
This observation is in line with \citet{hu-etal-2020}, who report a dissociation between perplexity and syntactic performance for English.

\begin{table}
\centering
\small
\begin{tabular}{c|rrr||c}
\hline
    & \multicolumn{1}{c}{flat}
    & \multicolumn{1}{c}{left}
    & \multicolumn{1}{c||}{right}
    & \\\hline
X   & 0.80$_{\pm.00}$ & 0.34$_{\pm.00}$ & 0.48$_{\pm.00}$ & \multirow{3}{*}{English}\\
POS & 0.79$_{\pm.00}$ & 0.57$_{\pm.00}$ & 0.70$_{\pm.00}$ & \\
DEP & \bf{0.82}$_{\pm.01}$ & 0.59$_{\pm.01}$ & 0.70$_{\pm.00}$ & \\\hline
X   & 0.79$_{\pm.00}$ & 0.37$_{\pm.00}$ & 0.58$_{\pm.00}$ & \multirow{3}{*}{French}\\
POS & \bf{0.86}$_{\pm.00}$ & 0.63$_{\pm.00}$ & 0.74$_{\pm.00}$ & \\
DEP & \bf{0.86}$_{\pm.01}$ & 0.65$_{\pm.01}$ & 0.75$_{\pm.00}$ & \\\hline
X   & 0.90$_{\pm.00}$ & 0.44$_{\pm.00}$ & 0.59$_{\pm.00}$ & \multirow{3}{*}{German}\\
POS & 0.85$_{\pm.00}$ & 0.74$_{\pm.00}$ & 0.76$_{\pm.00}$ & \\
DEP & \bf{0.91}$_{\pm.08}$ & 0.76$_{\pm.00}$ & 0.77$_{\pm.00}$ & \\\hline
X   & 0.81$_{\pm.01}$ & 0.41$_{\pm.00}$ & 0.58$_{\pm.00}$ & \multirow{3}{*}{Hebrew}\\
POS & \bf{0.83}$_{\pm.00}$ & 0.65$_{\pm.00}$ & 0.73$_{\pm.00}$ & \\
DEP & \bf{0.83}$_{\pm.00}$ & 0.65$_{\pm.00}$ & 0.72$_{\pm.00}$ & \\\hline
X   & 0.80$_{\pm.00}$ & 0.41$_{\pm.00}$ & 0.59$_{\pm.00}$ & \multirow{3}{*}{Russian}\\
POS & \bf{0.83}$_{\pm.00}$ & 0.62$_{\pm.00}$ & 0.73$_{\pm.00}$ & \\
DEP & 0.82$_{\pm.01}$ & 0.58$_{\pm.00}$ & 0.68$_{\pm.00}$ & \\
\hline
\end{tabular}
\caption{F1 score of predicted CTree. We regard a resulting CTree of each conversion as a gold tree.}
\label{tab:10M-f1}
\end{table}
\begin{figure}[t]
  \includegraphics[width=\linewidth]{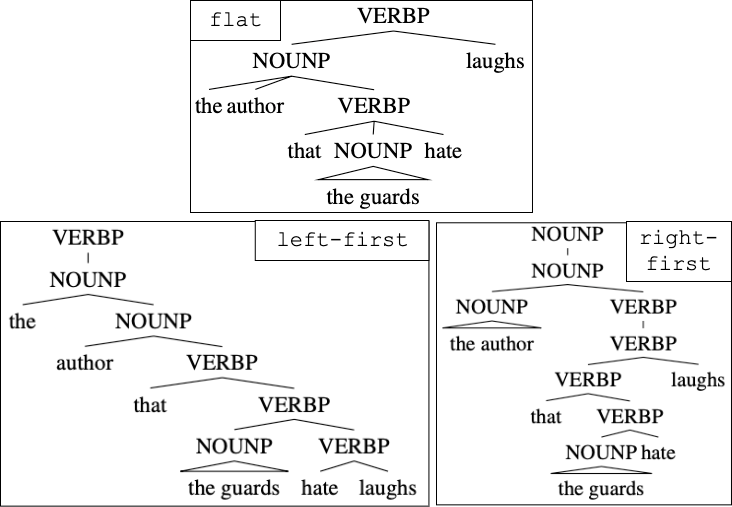}
  \caption{Structures of a CLAMS example predicted by \{flat, left-first, right-first\}-POS RNNG. This example is solvable only by flat-POS RNNG across all seeds.}
  \label{fig:predtree}
\end{figure}
\subsection{Why Does Flat Structure Perform Well?}
\label{ssec:f1}
As one possible reason why flat structure is optimal among the three structures presented, we conjecture that the parseability of the structure is involved.
To test this hypothesis, we calculated the F1 score between the gold CTrees of the test set and the structures predicted by RNNG for each setting.
Table~\ref{tab:10M-f1} shows the result.
The tendencies of F1 scores are consistent across languages: 1) Flat structures show highest F1 score. 2) While scores of flat structures are stable regardless of their labelings, the rest two structures exhibit lower score on X-label.
As a whole, the result reflects the tendency discussed in Section~\ref{ssec:10M-result}, which supports our hypothesis.

To further investigate the link between parseability and the capability of solving the task, we obtained parse trees of CLAMS examples that are solvable only by flat RNNG across all seeds.
We found that only flat RNNG produces a correct constituency tree, and structures obtained from left-first and right-first RNNGs are incorrect on a critical point.
For example, in Figure~\ref{fig:predtree}, while the relation between the subject ``author'' and the target verb ``laughs''  is analyzed clearly in the flat structure, it is ambiguous in the rest, possibly causing the misinterpretation that the subject is ``guards''.

These findings indicate the importance of choosing the correct tree structure for syntax-aware language modeling; it should be not only hierarchical, but also as parseable as possible.

Through analysis of the conversions, we found that (1) flat structure performs stably well in every setting. (2) while CLAMS accuracy of flat structure does not differ significantly depending on its labeling, for perplexity, flat-DEP performs the best for more than half of the languages and no inferiority can be observed for the other languages.
Therefore, we conclude that \emph{flat-DEP} conversion is the most robust conversion among languages.

\section{Advantage of Syntax Injection to LMs in a Multilingual Setting}
\label{sec:flatDEP}
In this section, we demonstrate the benefits of injecting syntactic biases into the model in a multilingual setting.
We obtained the CLAMS score of RNNG trained on the flat-DEP treebank (\emph{flat-DEP RNNG} for short) and compared it against baselines.

\begin{table*}
\centering
\scriptsize
\begin{tabular}{c|p{1cm}p{1cm}p{1cm}p{1cm}p{1cm}p{1cm}p{1cm}p{1cm}||c}
\hline
    & Simple
    & VP coord (short)
    & VP coord (long)
    & Across subj. rel.
    & Within obj rel.
    & Across obj rel.
    & Across prep.
    & Average
    & \\\hline
flat-DEP RNNG & 0.99$_{\pm.01}$ & 0.87$_{\pm.02}$ & 0.91$_{\pm.04}$ & 0.95$_{\pm.02}$ & 0.92$_{\pm.05}$ & 0.92$_{\pm.06}$ & 0.93$_{\pm.04}$ & {\bf 0.93}$_{\pm.02}$ & \multirow{4}{*}{English}\\
LSTM (N20) & 0.93$_{\pm.03}$ & 0.85$_{\pm.01}$ & 0.83$_{\pm.04}$ & 0.85$_{\pm.04}$ & 0.83$_{\pm.05}$ & 0.77$_{\pm.04}$ & 0.87$_{\pm.02}$ & 0.85$_{\pm.02}$ & \\
LSTM (M20) & 1.00$_{\pm.00}$ & 0.94$_{\pm.01}$ & 0.76$_{\pm.06}$ & 0.60$_{\pm.06}$ & 0.89$_{\pm.01}$ & 0.55$_{\pm.05}$ & 0.63$_{\pm.02}$ & 0.77$_{\pm.03}$  & \\
mBERT (M20) & 1.00 & 1.00 & 0.92 & 0.88 & 0.83 & 0.87 & 0.92 & 0.92 & \\\hline
flat-DEP RNNG & 1.00$_{\pm.00}$ & 1.00$_{\pm.00}$ & 1.00$_{\pm.00}$ & 1.00$_{\pm.00}$ & 1.00$_{\pm.00}$ & 1.00$_{\pm.00}$ & 1.00$_{\pm.00}$ & {\bf 1.00}$_{\pm.00}$ & \multirow{4}{*}{French}\\
LSTM (N20) & 1.00$_{\pm.00}$ & 1.00$_{\pm.00}$ & 0.97$_{\pm.03}$ & 0.92$_{\pm.06}$ & 0.85$_{\pm.03}$ & 0.75$_{\pm.01}$ & 1.00$_{\pm.00}$ & 0.93$_{\pm.01}$ & \\
LSTM (M20) & 1.00$_{\pm.00}$ & 0.97$_{\pm.01}$ & 0.85$_{\pm.05}$ & 0.71$_{\pm.05}$ & 0.99$_{\pm.01}$ & 0.52$_{\pm.01}$ & 0.74$_{\pm.02}$ & 0.83$_{\pm.02}$  & \\
mBERT (M20) & 1.00 & 1.00 & 0.98 & 0.57 & --- & 0.86 & 0.57 & 0.83 & \\\hline
flat-DEP RNNG & 1.00$_{\pm.00}$ & 0.99$_{\pm.01}$ & 0.98$_{\pm.01}$ & 1.00$_{\pm.00}$ & 0.88$_{\pm.04}$ & 0.99$_{\pm.01}$ & 0.97$_{\pm.02}$ & {\bf 0.97}$_{\pm.01}$ & \multirow{4}{*}{German}\\
LSTM (N20) & 0.99$_{\pm.01}$ & 0.97$_{\pm.03}$ & 0.92$_{\pm.05}$ & 0.99$_{\pm.01}$ & 0.72$_{\pm.01}$ & 0.97$_{\pm.02}$ & 0.94$_{\pm.01}$ & 0.93$_{\pm.01}$ & \\
LSTM (M20) & 1.00$_{\pm.00}$ & 0.99$_{\pm.02}$ & 0.96$_{\pm.04}$ & 0.94$_{\pm.04}$ & 0.74$_{\pm.03}$ & 0.81$_{\pm.09}$ & 0.89$_{\pm.06}$ & 0.90$_{\pm.04}$  & \\
mBERT (M20) & 0.95 & 0.97 & 1.00 & 0.73 & --- & 0.93 & 0.95 & 0.92 & \\\hline
flat-DEP RNNG & 0.97$_{\pm.01}$ & 0.99$_{\pm.00}$ & 0.92$_{\pm.03}$ & 0.95$_{\pm.02}$ & 1.00$_{\pm.00}$ & 0.84$_{\pm.05}$ & 0.95$_{\pm.01}$ & {\bf 0.95}$_{\pm.01}$ & \multirow{4}{*}{Hebrew}\\
LSTM (N20) & 0.97$_{\pm.00}$ & 0.95$_{\pm.04}$ & 0.85$_{\pm.02}$ & 0.89$_{\pm.02}$ & 0.94$_{\pm.01}$ & 0.63$_{\pm.04}$ & 0.93$_{\pm.01}$ & 0.88$_{\pm.00}$ & \\
LSTM (M20) & 0.95$_{\pm.01}$ & 1.00$_{\pm.01}$ & 0.84$_{\pm.06}$ & 0.91$_{\pm.03}$ & 1.00$_{\pm.01}$ & 0.56$_{\pm.01}$ & 0.88$_{\pm.03}$ & 0.88$_{\pm.02}$  & \\
mBERT (M20) & 0.70 & 0.91 & 0.73 & 0.61 & --- & 0.55 & 0.62 & 0.69 & \\\hline
flat-DEP RNNG & 0.89$_{\pm.02}$ & 0.94$_{\pm.02}$ & 1.00$_{\pm.00}$ & 0.93$_{\pm.00}$ & 0.99$_{\pm.01}$ & 0.92$_{\pm.02}$ & 0.85$_{\pm.03}$ & {\bf 0.93}$_{\pm.01}$ & \multirow{4}{*}{Russian}\\
LSTM (N20) & 0.91$_{\pm.01}$ & 0.97$_{\pm.00}$ & 0.97$_{\pm.02}$ & 0.98$_{\pm.00}$ & 0.90$_{\pm.04}$ & 0.85$_{\pm.07}$ & 0.86$_{\pm.02}$ & 0.92$_{\pm.01}$ & \\
LSTM (M20) & 0.91$_{\pm.01}$ & 0.98$_{\pm.02}$ & 0.86$_{\pm.04}$ & 0.88$_{\pm.03}$ & 0.95$_{\pm.04}$ & 0.60$_{\pm.03}$ & 0.76$_{\pm.02}$ & 0.85$_{\pm.03}$  & \\
mBERT (M20) & 0.65 & 0.80 & --- & 0.70 & --- & 0.67 & 0.56 & 0.68 & \\\hline
\end{tabular}
\caption{CLAMS scores for flat-DEP RNNG and baselines. LSTM (N20) is a model of which hyperparameters are set as with \citet{noji-takamura-2020}. LSTM (M20) and mBERT (M20) scores are quoted from Table 1, 2 and 5 in \citet{mueller-etal-2020}. Hyphen means that all focus verb for the corresponding setting were out-of-vocabulary.}
\label{tab:80M-clams}
\end{table*}
\paragraph{Experimental setup}
The experiment was conducted in as close setting to the previous work as possible.
Following \citet{mueller-etal-2020}, we extracted Wikipedia articles of 80M tokens as training set.
The hyperparameters of LSTM LM are set following \citet{noji-takamura-2020} because it performs the best for the dataset of \citet{marvin-linzen-2018}\footnote{Since English set of CLAMS is a subset of \citet{marvin-linzen-2018}, it is reasonable to choose this model to validate the multilingual extendability.}.
We used subword units with a vocabulary size of 30K, and the sizes of RNNG and LSTM LM are set to be the same (35M).

\paragraph{Result}
Table~\ref{tab:80M-clams} shows the result.
In addition to scores from the models we trained (flat-DEP RNNG, LSTM (N20)), we display scores of LSTM LM and mBERT reported in the original paper (LSTM (M20) and mBERT (M20), \citealp{mueller-etal-2020}).
Overall, we can see the superiority of RNNG across languages, especially for the tasks that require analysis on long distance dependency; `VP coord (long)', `Across subj. rel.', `Across obj. rel.', and `Across prep'.
While previous work suggested that LSTM LMs potentially have a limitation in handling object relative clauses
\citep{noji-takamura-2020}, our result suggests that RNNG does not have such a limitation thanks to explicitly injected syntactic biases.
% However, we also observe exceptions in German and Russian.
% For German, we can see a clear valley; the accuracy for Within obj. rel. is unnaturally low.
% This outlier-like tendency was also pointed out in \citet{mueller-etal-2020}, and they describe the reason as its marking of cases, a unique feature among five languages.
% TODO: This is also the case in our situation (or may be no because of subword. we need some investigation).
% For Russian, LSTM LM performs slightly better for the easy tasks (Simple, VP coord (short)).

\section{Discussion}
\label{sec:discussion}
We discussed the CTree structure that works robustly regardless of the language and the superiority of injecting syntactic bias to the model.
Our claim is that we can construct language-independent syntax-aware LMs by seeking the best structure for learning RNNGs, which is backed up by our experiments based on five languages.
To make this claim firm, more investigations are needed from two aspects: \textbf{fine-grained syntactic evaluation} and \textbf{experiment on typologically diverse languages}.

\paragraph{Fine-grained syntactic evaluation}
The linguistic phenomenon covered in CLAMS is only an agreement.
However, previous works have invented evaluation sets that examine more diverse syntactic phenomena for English (\citealp{hu-etal-2020}, \citealp{warstadt-etal-2020}).
We need such a fine-grained evaluation even in a multilingual setting, as superiority in agreement does not imply superiority in every syntactic knowledge; \citet{kuncoro-etal-2019} suggested that RNNG performs poorer than LSTM LM in capturing sentential complement or simple negative polarity items.
It is challenging to design a multiliugnal syntactic test set because even an agreement based on grammatical categories is not a universal phenomenon.
It is required to seek reasonable metrics that cover broad syntactic phenomena and are applicable to many languages.

\paragraph{Experiment on typologically diverse languages}
Languages included in CLAMS (English, French, German, Hebrew and Russian) are actually not typologically diverse.
Apart from language-specific features, all of them take the same ordering of (1) subject, verb, and object (SVO) (2) relative clause and noun (Noun-Relative clause) (3) adposition and noun phrase (preposition), and so on\footnote{Typological information is obtained from WALS: \url{https://wals.info/}}.
If we run the same experiment for a typologically different language, the result could be somewhat different.
Although some previous work focused on syntactic assessment of other languages (\citealp{ravfogel-etal-2018}; \citealp{gulordava-etal-2018}), such attempts are scarce.
As future work, it is needed to design an evaluation set based on other languages and explore the extendability to more diverse languages. 

\section{Conclusion}
\label{sec:conclusion}
In this paper, we propose a methodology to learn multilingual RNNG through dependency tree conversion.
We performed multiple conversions to seek the robust structure which works well multilingually, discussing the effect of multiple structures.
We demonstrated the superiority of our model over baselines in capturing syntax in a multilingual setting.
Since our research is the first step for multilingual syntax-aware LMs, it is necessary to conduct experiments on more diverse languages to seek a better structure.
We believe that this research would contribute to the field of theoretical/cognitive linguistics as well because an ultimate goal of linguistics is finding the universal rule of natural language.
Finding a reasonable structure in engineering would yield useful knowledge for that purpose.

\section*{Acknowledgements}
This paper is based on results obtained from a project JPNP20006, commissioned by the New Energy and Industrial Technology Development Organization (NEDO). For experiments, computational resource of AI Bridging Cloud Infrastructure (ABCI) provided by National Institute of Advanced Industrial Science and Technology (AIST) was used.

\bibliography{acl}
\bibliographystyle{acl_natbib}

%\appendix

%\section{Example Appendix}
%\label{sec:appendix}

%This is an appendix.

\end{document}